\pdfoutput=1

\documentclass[11pt]{article}

\usepackage{emnlp2021}

\usepackage{times}
\usepackage{latexsym}

\usepackage[T1]{fontenc}

\usepackage[utf8]{inputenc}

\usepackage{microtype}

\usepackage{amssymb}
\usepackage{pifont}
\newcommand{\cmark}{\ding{51}}%
\newcommand{\xmark}{\ding{55}}%
\usepackage{graphicx}
\usepackage{booktabs}
\usepackage{comment}

\usepackage{url}
\usepackage{hyperref}
\usepackage{booktabs}

\usepackage{microtype}
\usepackage{booktabs}
\usepackage{amsmath}
\usepackage{flexisym}
\usepackage{dcolumn}
\usepackage{kotex}
\usepackage{hhline}
\usepackage{verbatim}
\usepackage{multirow}
\usepackage{amssymb}
\usepackage{rotating}
\usepackage{subfigure}
\usepackage{dblfloatfix}
\usepackage{enumitem}
\usepackage{framed}
\usepackage[utf8]{inputenc}
\usepackage[english]{babel}
\usepackage{xcolor}
\usepackage{tabulary}
\usepackage{color}
\usepackage{arydshln}

\newcolumntype{L}[1]{>{\raggedright\let\newline\\\arraybackslash\hspace{0pt}}m{#1}}
\newcolumntype{C}[1]{>{\centering\let\newline\\\arraybackslash\hspace{0pt}}m{#1}}
\newcolumntype{R}[1]{>{\raggedleft\let\newline\\\arraybackslash\hspace{0pt}}m{#1}}

\title{QACE: Asking Questions to Evaluate an Image Caption}

\author{Hwanhee Lee$^{1}$, Thomas Scialom$^{2, 3}$, Seunghyun Yoon$^{4}$, Franck Dernoncourt$^{4}$, Kyomin Jung$^{1}$ \\
$^{1}$Dept. of Electrical and Computer Engineering, Seoul National University, Seoul, Korea \\
$^{2}$Sorbonne Université, CNRS, LIP6, F-75005 Paris, France\\
$^{3}$reciTAL, Paris, France, $^{4}$Adobe Research, San Jose, CA, USA\\
{\tt \{wanted1007, kjung\}@snu.ac.kr, thomas@recital.ai} \\
{\tt \{syoon, franck.dernoncourt\}@adobe.com} \\
}

\begin{document}

\maketitle
\begin{abstract}
In this paper, we propose QACE, a new metric 
based on \textbf{Q}uestion \textbf{A}nswering for \textbf{C}aption \textbf{E}valuation.
QACE generates questions on the evaluated caption and checks its content by asking the questions on either the reference caption or the source image.
We first develop QACE\textsubscript{Ref} that compares the answers of the evaluated caption to its reference, and report competitive results with the state-of-the-art metrics. To go further, we propose QACE\textsubscript{Img}, which asks the questions directly on the image, instead of reference. A Visual-QA system is necessary for QACE\textsubscript{Img}. Unfortunately, the standard VQA models are framed as a classification among only a few thousand categories. Instead, we propose Visual-T5, an \emph{abstractive} VQA system. The resulting metric, QACE\textsubscript{Img} is multi-modal, reference-less, and explainable. Our experiments show that QACE\textsubscript{Img} compares favorably w.r.t. other reference-less metrics. We will release the pre-trained models to compute QACE.\footnote{https://github.com/hwanheelee1993/QACE}
\end{abstract}

\section{Introduction}
\begin{figure*}[t]
\small
\centering
\includegraphics[width=1.75\columnwidth]{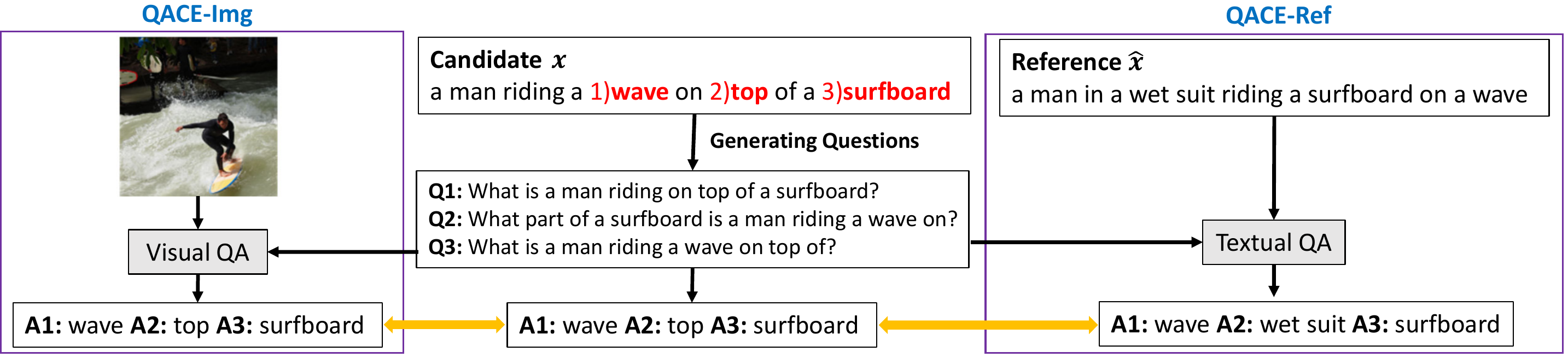}
\caption{
The overall flow of QACE. QACE extracts possible answer spans and generates answer-aware questions for a given candidate caption $x$. The VQA and TQA answer these questions given the image and reference captions, respectively. The correctness of the candidate caption is evaluated by comparing the answers.
}
\label{fig_QACE}
\end{figure*}

Image captioning is a task that aims to generate a description containing the main content of a given image. The field of caption generation is prolific~\cite{vinyals2015show, anderson2018bottom}, and it is, therefore, important to provide reliable evaluation metrics to compare the systems. Most of the prior works still report n-gram similarity metrics such as BLEU~\cite{papineni2002bleu} or CIDEr~\cite{vedantam2015cider}. However, these n-gram similarity metrics often fail to capture the semantic errors in the generated captions \cite{novikova-etal-2017-need}.
\\ \indent To overcome this limitation, we propose QACE, a radically different evaluation framework from n-gram metrics. QACE first generates questions about the candidate caption, and then checks if the answers are consistent w.r.t. either the reference or the source image. We depict QACE in Figure~\ref{fig_QACE}.
\\ \indent Specifically, we propose two variants of QACE, depending on what content the evaluated caption is compared to: QACE\textsubscript{Ref} when it is compared to the reference, and QACE\textsubscript{Img} when it is compared to the source image. QACE\textsubscript{Img} has the desired feature to be \emph{reference-less}, i.e., the score can be computed without requiring a gold reference. 
\\ \indent In this reference-less setup, a Visual Question Answering (VQA) system is required to answer those questions. However, in the VQA literature~\cite{antol2015vqa}, the task is usually seen as a classification task on 3k pre-defined answer choices (e.g., blue, sea, or banana). Therefore, these VQA models are not general QA systems; their usage off-the-shelf for QACE\textsubscript{Img} would limit the comparison to these very few pre-defined categories, which is not satisfying. To solve this issue, we also propose an abstractive VQA system Visual-T5 as a new module for QACE\textsubscript{Img} that can generate free-form abstractive answers given a textual question and an image. We conduct a human evaluation of Visual-T5 and show that it is capable of generating accurate abstractive answers. Using Visual-T5, we are now able to compare the answers of the candidate caption directly with the answers of the corresponding image. 
\\ \indent Experimental results show that our proposed QACE\textsubscript{Ref} and QACE\textsubscript{Img} show promising results compared to other reference and reference-less metrics on three benchmark datasets: Pascal50s~\cite{vedantam2015cider}, Composite~\cite{aditya2015images} and Flickr8k~\cite{hodosh2013framing}. Also, as shown in Figure~\ref{fig_QACE}, QACE has a natural form of interpretability through the visualization of the questions and the answers.

\section{Related Work}
\paragraph{Image Captioning Metrics}
Similar to other text generation tasks such as machine translation and summarization, n-gram similarity metrics such as BLEU, METEOR~\cite{banerjee2005meteor} and ROUGE~\cite{lin2004rouge} are arguably the standard in automatic evaluation. Among them, the most widely used metric is CIDEr~\cite{vedantam2015cider} which uses TF-IDF based weighted n-gram similarity. SPICE~\cite{anderson2016spice} metric is based on scene graph, while more recently, BERTScore~\cite{zhang2019bertscore} compute the similarity of the contextualized embeddings.   
Different from prior works, we are the first to use Question Generation (QG) and Question Answering (QA) to evaluate the image captions.

\paragraph{Question and Answering for Evaluation}
\citet{fisch2020capwap} proposes a new method to generate informal captions that can answer the visual questions. In our work, we focus on caption evaluation using the QA systems, not on generating the captions.
Several QA-based evaluation metrics~\cite{ scialom2019answers, wang2020asking} are recently proposed to evaluate abstractive summarization. However, all those prior works are limited to text-to-text evaluation, while our work develops a multi-modal metric. 

\section{QACE}
We propose QACE, which is a QG- and QA-based framework for evaluating an image caption. As shown in Figure~\ref{fig_QACE}, QACE first extracts answer candidates (i.e., 1) wave, 2) top, 3) surfboard) from a candidate caption and generates corresponding questions. With these questions, visual-QA (VQA) and textual-QA (TQA) models answers given their context (i.e., image and reference $\hat{x}$). By comparing the answers from each source, we can directly judge the correctness of the candidate caption. 

\subsection{Question Generation}
\label{ssec:question_generation}
The goal of this component is to generate questions that ask the primary information of the candidate caption. Our QG model is a text-to-text generation model (i.e., T5~\cite{raffel2020exploring}), fine-tuned on SQuAD v2 \cite{rajpurkar2018know} to generate answer-aware questions.
Given a caption, we extract possible answer span; in particular, we focus on extracting noun phrases since they mostly contain salient information and can be easily foiled~\cite{shekhar2017foil}. 
We argue that questions generated on this salient information should be answered similarly from the image or the captions if they share the same information.

\subsection{Question Answering}
For QACE\textsubscript{Ref}, we use a TQA model. We train T5 to answer the generated questions (see~\ref{ssec:question_generation}) with the reference captions as context. 
Conversely, QACE\textsubscript{Img} requires a VQA model. We propose a new architecture, Visual-T5, that can generate abstractive answers given an image and a question, as opposed to the standard multiple-choice VQA.

\begin{figure}[t]
\small
\centering
\includegraphics[width=0.95\columnwidth]{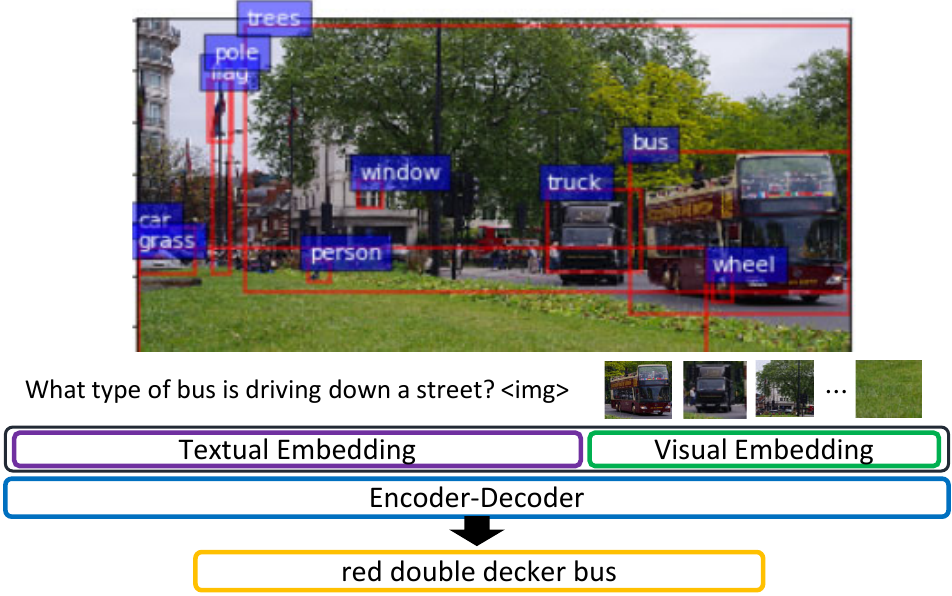}
\caption{
The overview of Visual-T5, an abstractive VQA model. We embed questions with additional special separation token and concatenate the visual embeddings to make inputs for T5. 
}
\vspace{-5mm}
\label{fig_visualt5}
\end{figure}
\subsection{Abstractive Visual Question Answering}
\label{ssec:absVQA}
When no reference captions are available, one of the most important parts of QACE is the VQA model that can produce correct answers.
To move beyond VQA as a classification task, we are the first, to the best of our knowledge, to develop an abstractive VQA model that can generate free-form answers. Specifically, we enable multimodal encoding for T5, inspired by the previous works on adapting pre-trained language models for multimodal tasks~\cite{scialom2020bert}. We illustrate our proposed Visual-T5 in Figure~\ref{fig_visualt5}. Based on default T5 architecture, Visual-T5 has an additional visual embedding layer that encodes regional features of the image from Faster RCNN~\cite{ren2015faster}. This linear layer maps detection features to 768 dimensions, same as the dimension of textual embedding. This 768d features are therefore considered as a standard token in Visual-T5, which can encode an image and a question together. We provide more details in Appendix. 

\subsection{QACE Metric}
\label{ssec:qace_metric}
For a given candidate caption $x$, We use $QG$ to generate questions $Q$\,{=}\,($q_1$, ..., $q_M$) for all of $M$ noun phrases of $x$. Then, we compare the answers for each question in $Q$ on $x$ with the answers on the reference source. We introduce two QACE variants, QACE\textsubscript{Ref} for which the reference caption is compared, and QACE\textsubscript{Img} for which the source image is compared. Using $QG$ and $QA$, we compute QACE\textsubscript{Ref} and QACE\textsubscript{Img} as follows:
\begin{equation}
\begin{aligned}
     \tiny{QACE} = \tiny{\frac{\Sigma_{i=1}^{M}f(QA(q_i, x), QA(q_i, ctx))}{M}},
\end{aligned}
\label{eq:qace}
\end{equation}
where $ctx$ corresponds to the image for QACE\textsubscript{Img} and the gold reference for QACE\textsubscript{Ref}, $f(A_1, A_2)$ is the function that measures the similarity between two answers $A_1$ and $A_2$. The standard metric in QA is the F1, as introduced by \citet{rajpurkar2016squad}. However, two abstractive answers can be similar but written in two different ways, limiting the effectiveness of a naive F1. Hence, in addition to the F1, we propose to use the BERTScore. Finally, we also complete the similarity metrics using the answerability of the questions for function $f$, in order to measure whether the question is answerable. The answerability corresponds to $1-P_{unanswerable}$, where $P_{unanswerable}$ is the probability attributed by the model to the token \emph{unanswerable}.\footnote{SQuAD v2 contains unanswerable questions, for which we associate the token \emph{unanswerable}  as the correct answer during training. Therefore, our QA model associates this token with the probability that the question is not answerable.} To consider all the different aspects, we use the average of three values computed using each function as the default value of QACE.

\section{Synthetic Data Generation for VQA}
As discussed in~\ref{ssec:absVQA}, relying on a VQA dataset such as VQA v2~\cite{goyal2017making} limits possible answers to a small size of pre-defined categories.
To train a general and abstractive VQA model, we create synthetic abstractive VQA datasets. We generate Questions/Answers pairs using the captions in the training set of MS-COCO~\cite{lin2014microsoft}.
Specifically, we extract noun phrases from a reference caption and generate an answer-aware question using our QG model. To increase the validity of these synthetic questions, we apply the round trip consistency~\cite{alberti2019synthetic}, filtering out the questions for which the QA model predicts a different answer than the extracted noun phrase. We convert these synthetic QA dataset to create \{\textit{question, answer, image}\} triples by concatenating the corresponding images to these captions.\\
\indent In addition, we randomly add 20\% of unanswerable questions\footnote{We consider an image and a question that are not paired to be unanswerable, and do negative sampling.} to the synthetic training set, so that the model learns to judge the answerability of a given question. 
Through this, if a candidate caption contains any hallucinating content that is not included in the image, questions about it can be marked as unanswerable by our VQA model, as shown in the second example of Figure~\ref{fig_case_study}. This synthetic dataset enables the training of the abstractive VQA model. We report the performance of the model through a human evaluation in Section~\ref{par_vqa_performance}.

\section{Experiments}
\subsection{Benchmark Dataset}
We evaluate our proposed metric on three benchmark datasets (i.e. human annotations), \textit{PASCAL-50S}, \textit{Composite} and \textit{Flickr8k}.

\indent \textit{PASCAL-50S} provides 4k caption triplet $<$\textit{A}, \textit{B}, \textit{C}$>$, where ''\textit{A}" is composed of 50 reference captions(\textit{A}) and two candidate captions(\textit{B}, \textit{C}) for the given image. There are human judgments as to which ``\textit{B}" or ``\textit{C}" is more appropriate caption for a given image compared to ``\textit{A}".

\indent \textit{Composite} is composed of 11,985 human judgments scores range from 1 to 5 depending on the relevance between each candidate caption-image pair with 5 reference captions.

\indent \textit{Flickr8k} provides three human-expert judgments for 5,822 candidate caption-image pairs. The scores are from 1 to 4, depending on the relevance of each caption-image pair.

\subsection{Results and Discussions}
\begin{table}[!t]
\centering 
\small
\resizebox{0.95\columnwidth}{!}{%
\begin{tabular}{lcccc}
\toprule

                  & Ref?   & Pascal50s  & Composite     & Flickr8k \\
\midrule
BLEU-4            & \cmark & 65.2       & 45.7          & 28.6 \\
ROUGE-L           & \cmark & 67.7       & 47.7          & 30.0 \\
METEOR            & \cmark & \textbf{80.5} & 46.6       & 40.3 \\ 
CIDEr             & \cmark & 77.8       & 47.4          & 41.9 \\
SPICE             & \cmark & 76.1       & 48.6          & \textbf{45.7} \\
BERTScore         & \cmark & 72.0       & 45.6          & 30.5 \\
QACE-Ref (ours)      & \cmark & 75.1       & 49.3 & 40.5 \\
\hspace{3mm} \textit{F1}             & \cmark    & 57.5          & \textbf{55.1}          & 9.2 \\
\hspace{3mm} \textit{BERTScore}      & \cmark    & 76.4          & 46.0          & 30.9  \\
\hspace{3mm} \textit{Answerability}  & \cmark    & 71.6 & 47.3          & 39.0   \\
\midrule
-Perplexity       & \xmark  & 46.8       & 1.7*           & 10.1  \\
VIFIDEL           &  \xmark & 69.0       & 13.1          & \textbf{33.6}  \\
QACE-Img (ours)                         & \xmark    & \textbf{70.0} & \textbf{19.1} & 29.1 \\
\hspace{3mm} \textit{F1}             & \xmark    & 62.0          & 12.5          & 27.3 \\
\hspace{3mm} \textit{BERTScore}      & \xmark    & 65.9          & 12.8          & 27.1  \\
\hspace{3mm} \textit{Answerability}  & \xmark    & \textbf{74.5} & 15.7          & 27.8   \\

\bottomrule
\end{tabular}
}
\caption{First column represents the accuracy of matches between human judgments in PASCAL50s. Columns 2 to 3 show the Kendall Correlation between human judgments and various metrics. All p-values in the results are $<$ 0.05 except for *. 
}
\vspace{-6mm}
\label{table_main_result}
\end{table}

\paragraph{Computation Details}
For all of the results on reference based metrics we reported in the paper, we compute the average of each metric score with each reference for all of the references on each dataset.

\paragraph{QACE Performance} 
We compare our proposed method with the following widely used metrics: BLEU-4, ROUGE-L~, METEOR, CIDEr, SPICE, and the BERTScore.
We present the experimental results for all three datasets in Table~\ref{table_main_result}. 
For the reference-aware metrics, QACE\textsubscript{Ref} shows best results on Composite and comparable to the best metrics for Pascal50s and Flickr8k, indicating the relevance of a QA based metric to evaluate image captioning.\\ 
\indent For the reference-less metrics, all the correlations are lower this time, showing the difficulty of evaluating the captions without reference. Nonetheless,  among these metrics, QACE\textsubscript{Img} shows the best results for Pascal50s and Composite and comparable results in Flickr8k. For Flickr8k, we found that more than half of the human judgments of the candidate captions are less than 0.2 as 0 to 1 scale. In other words, most of the captions in this dataset are totally not related to the image. For this reason, most of the generated questions are unanswerable for an image and we explain that this leads to relatively lower performance of QACE\textsubscript{Img} in Flickr8k compared to other metrics.\\
\indent Furthermore, We investigate the independent contribution of each answer similarity function, $f$, in computing QACE and present the results in Table~\ref{table_main_result} 
(note that default QACE-Img uses the mean of F1, BERTScore and answerability). 
The table reveals that each similarity function has a different aspect, and averaging three results suggests the best performance for two of three datasets. 

\paragraph{VQA Model Performance}

Visual-T5 is one of the main components of QACE\textsubscript{Img}. 
Since it can generate free-form answers, its automatic evaluation is challenging. We therefore conduct a human evaluation on 200 examples randomly sampled from the test set. We hire three annotators to judge whether the generated answer is correct or not given the image. On the majority vote from three annotators, VQA model correctly answers for the average 69\% of the examples.
Among these 69\% correct answers, half of them were written differently from the original answer, showing that our model can generate abstractive answers.
\label{par_vqa_performance}

\paragraph{Case study}
\begin{figure}[t]
\small
\centering
\includegraphics[width=1.0\columnwidth]{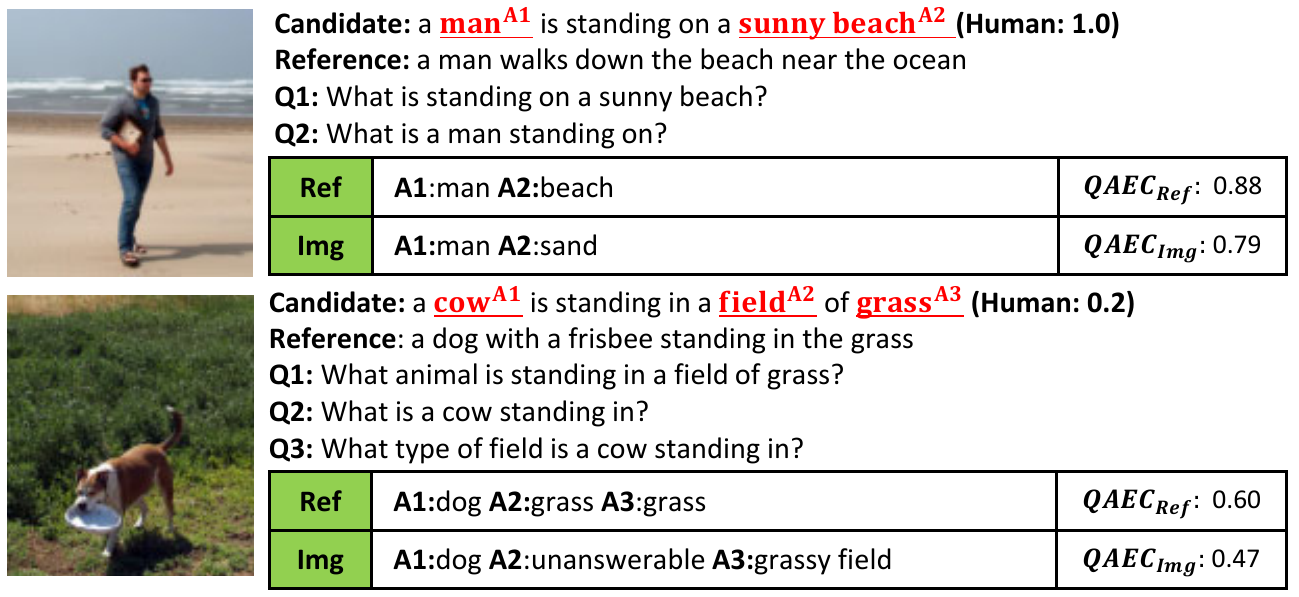}
\caption{
Case study on QACE metric. Human judgments are normalized to between 0 and 1.
}
\vspace{-5mm}
\label{fig_case_study}
\end{figure}

Different from the previous metrics, QACE can be easily interpreted through the visualization of the generated questions and the following answers as shown in Figure~\ref{fig_case_study}. In the first example, we observe that the second question is answered differently by the VQA model (\textit{sand} VS  \textit{beach}). Despite, the answer itself being correct - it is true that the man is standing on the sand - it results in a lower score for QACE\textsubscript{Img} compared to QACE\textsubscript{Ref}. This emphasizes the importance to use other similarity metrics than the F1 when comparing two answers (see Section \ref{ssec:qace_metric}). For instance, BERTScore should be able to consider closer \textit{sand} and \textit{beach} than \textit{sand} and a random word. \\
\indent The second example is very illustrative: 
for the first question, both TQA and VQA answer \textit{dog}, hence detecting an error in the candidate caption that talks about a \textit{cow}. 
The second question refers to the \textit{cow}, which makes it ambiguous. The VQA model considers it as \textit{unanswerable}, while the TQA model correctly answers \textit{grass}. 
Following this study, we expect that QACE\textsubscript{Img} can be improved through 
a finer answer comparison method in future work.

\section{Conclusion}
In this paper, we propose QACE, a captioning metric that directly compares each content in the candidate caption with either the source image or a gold reference caption by asking questions. 
To enable asking questions directly on the source image, we introduce Visual T5, an abstractive VQA model to generate free-form visual answers, for which we report strong results based on a human evaluation.
Our proposed metric can be applied in both reference and reference-less settings. It holds high explainability and compares favorably to the state-of-the-art in terms of correlations with human judgments.


\section*{Acknowledgements}
We thank anonymous reviewers for their constructive and meaningful comments. K. Jung is with ASRI, Seoul National University, Korea. This work was supported by the National Research Foundation of Korea (NRF) grant funded by the Korea government (No. 2021R1A2C2008855). This work was partially funded by gifts from Adobe Research.

\section*{Ethical Considerations}
We compensate the workers with competitive pay, which is above hourly USD \$10 for the human evaluation of VQA model. Furthermore, we used public datasets to train the models.


\bibliographystyle{acl_natbib}
\bibliography{emnlp2021}

\clearpage

\appendix
\section{Experimental Details}

\subsection{Reproducibility Checklist}

\paragraph{Source Code}
We attach the source for computing QACE and training Visual-T5. In the question generation components, we use the Noun Chunks extractor from spaCy.\footnote{https://spacy.io/usage/linguistic-features\#noun-chunks}

\paragraph{Computing Infrastructure}
We use AMD Ryzen Threadripper 2950X (3.50 GHz) with GeForce RTX 2080 Ti for the experiments. The software environments are Python 3.6.6 and PyTorch 1.3.1.

\paragraph{Average runtime for each approach}
It takes average one second to generate all questions for a given candidate caption using a pre-trained question generation model. And it takes average about 0.1 seconds to compute visual and textual answers, and comparing the answers. For training VQA model, Visual-T5, each epoch takes 40 minutes using a single RTX 2080 Ti GPU.

\paragraph{Hyperparameters}
We use the pre-trained \textit{t5-base} for question generation and TQA model in the model repository\footnote{https://github.com/mrm8488/question_generation} of huggingface~\cite{wolf2020transformers}.
We use \textit{t5-small} to fine-tune our VQA model. Based on \textit{t5-small}, we added single linear layer to encode visual features and then train the model for 5 epochs with batch size of 128. The number of the synthetic training set is 1 million and we split the dataset into 9:1 proportion for training and validation. For the max sequence length, we set 64 to the input sequence including the visual tokens, and set 32 to output sequence.

\paragraph{Number of Model Parameters} 
The number of parameters for QG is 222.9M, TQA is 222.9M and VQA is 61.6M.

\subsection{Significance Test}
We conduct a standard way to test the significance of the correlation coefficient for all of the reported correlation coefficients in the paper. We use a t-test that uses a null hypothesis, which is an absence of association, and report the p-value for each coefficient.

\section{Abstractive Visual Question Answering}
\label{app:model}

We provide the training details including the additional output examples of our proposed abstractive VQA model, Visual-T5 in this section.

\subsection{Visual Embedding}
We extract the regional features for each object using Faster RCNN~\cite{ren2015faster}. We fixed the number of boxes to 36 and each regional feature consists of dimension 2048 and 6 additional dimensions consists of the location and the size of each box. We concatenate this additional dimensions to make dimension of 2054 for each regional feature. And single linear layer maps these 2054d features to 768d to be considered as a token in T5.

\subsection{Answer Examples}
\begin{figure}[t]
\small
\centering
\includegraphics[width=0.95\columnwidth]{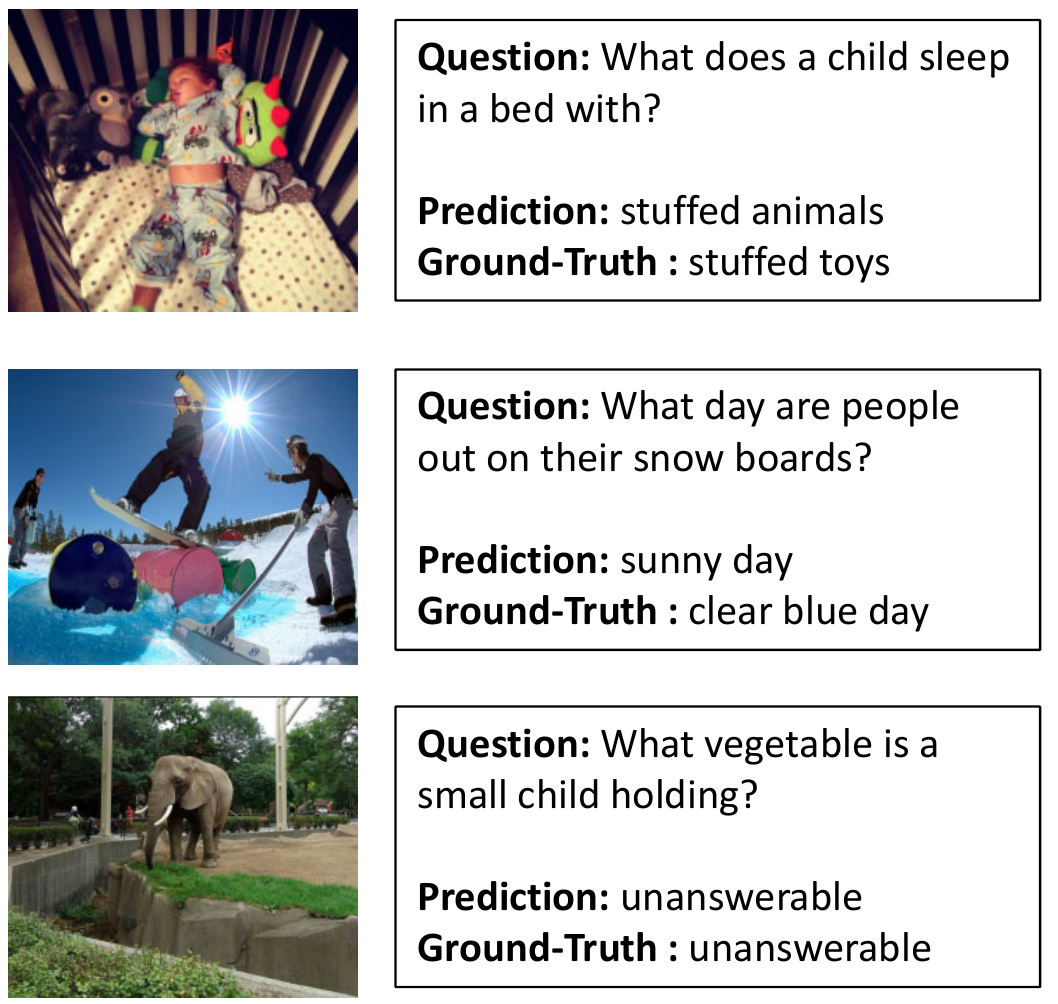}
\caption{
Various output examples on the evaluation set of abstractive VQA model, Visual-T5.
}
\label{fig_vqa_examples}
\end{figure}
We provide more examples of our abstractive VQA models in Figure~\ref{fig_vqa_examples}. We observe that many predicted answers are correct, but expressed in a different form as in the first and the second example. Also, model outputs \emph{unanswerable} to the questions that are unanswerable for a given image like the third example.

\subsection{Answerability}
We make unanswerable visual questions by randomly sampling the questions from the different images to the given image. We mixed 20\% of these unanswerable questions similar to the third example in Figure~\ref{fig_vqa_examples} to train VQA model.

\subsection{Human Evaluation}
\begin{figure}[!t]
\small
\centering
\includegraphics[width=0.9\columnwidth]{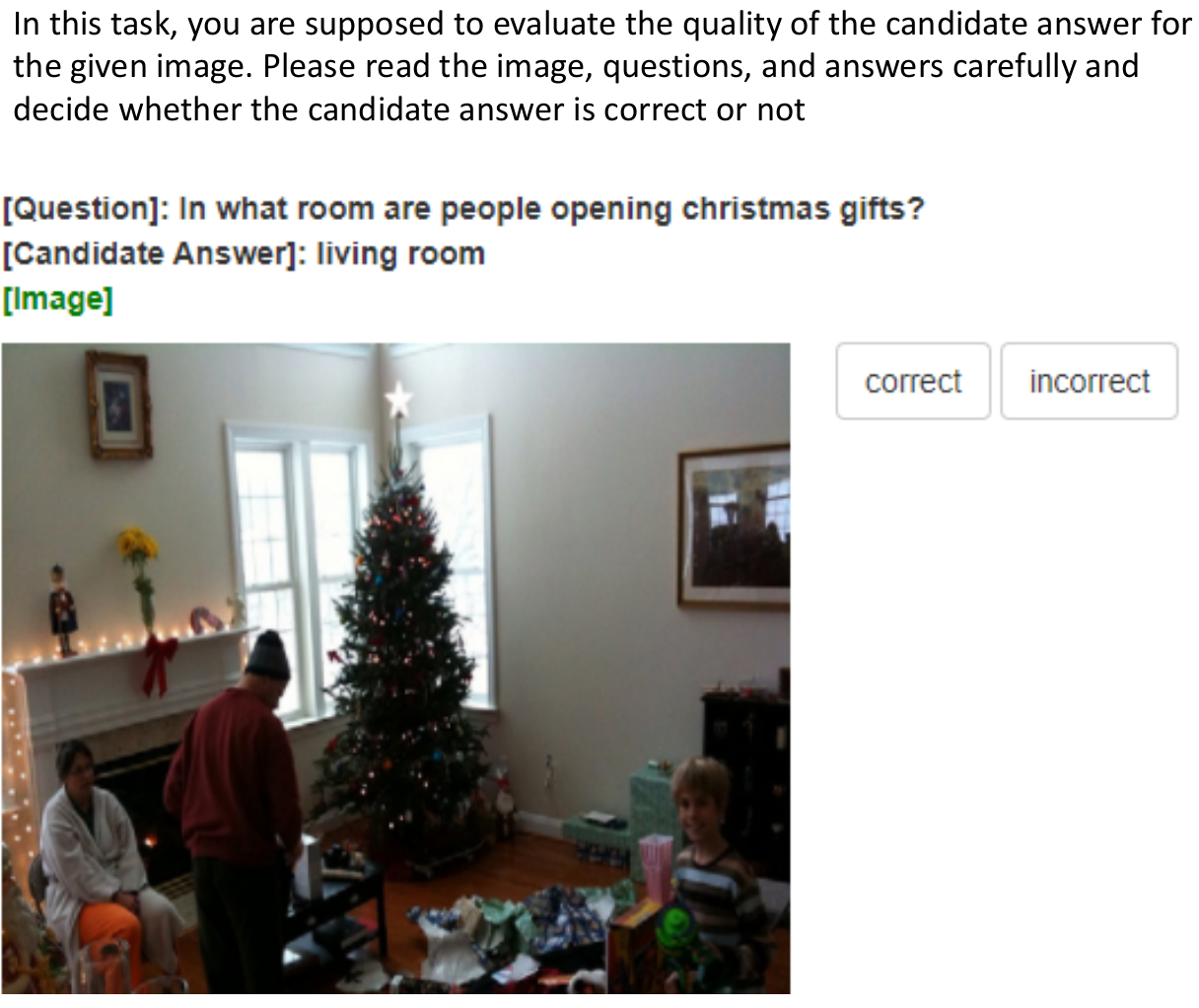}
\caption{
Full instructions and interface to workers for evaluating the answers of VQA model.
}
\label{fig_instruction}
\end{figure}
We hire the workers whose locations in one of the US, UK, CA, NZ, AU to guarantee the fluency in English. We restrict the workers whose HIT approval rates are higher than 95\%, and minimum hits are over 500. We pay workers more than USD \$10 in an hour through several preliminary experiments on the compensation.
We provide the full instructions and the interface in Figure~\ref{fig_instruction}.
We compute the annotator agreement using Krippendorff's $\alpha$~\cite{krippendorff1970estimating}. We observe that Krippendorff's $\alpha$ is 0.56 that indicates a ``moderate`` agreement according to one of the referenced guidelines~\cite{landis1977measurement} for kappa-like measures.


\end{document}